%% file: acl_latex.tex
\pdfoutput=1

\documentclass[11pt]{article}
\usepackage[table]{xcolor}
\definecolor{verylightgray}{gray}{0.82}
\usepackage[]{acl}

\usepackage{times}
\usepackage{latexsym}
\usepackage{amsmath}
\usepackage{multirow}
\usepackage{adjustbox}
\usepackage{booktabs} 
\usepackage{subcaption}
\usepackage[T1]{fontenc}
\usepackage[utf8]{inputenc}
\usepackage{microtype}
\usepackage{inconsolata}
\usepackage{graphicx}
\usepackage[most]{tcolorbox}

\newcommand{\todo}[1]{{\color{red} #1}}

\title{Budget-Aware Anytime Reasoning with LLM-Synthesized Preference Data}

\author{
 \textbf{Xuanming Zhang\textsuperscript{1}\thanks{Work done during an internship at Oracle.}},
 \textbf{Shwan Ashrafi\textsuperscript{2}},
 \textbf{Aziza Mirsaidova\textsuperscript{2}},
 \textbf{Amir H. Rezaeian\textsuperscript{2}}, \\
 \textbf{Miguel Ballesteros\textsuperscript{2}},
 \textbf{Lydia B. Chilton\textsuperscript{1}},
 \textbf{Zhou Yu\textsuperscript{1}},
 \textbf{Dan Roth\textsuperscript{2}}
\\
\\
 \textsuperscript{1}Columbia University,
 \textsuperscript{2}Oracle AI
\\
\small{
 \texttt{
   xz2995@columbia.edu, \{shwan.ashrafi,dan.roth\}@oracle.com
 }}
}

\begin{document}
\maketitle
\begin{abstract}
\input{sections/0-abstract}
\end{abstract}

\tcbset{
    promptbox/.style={
        colback=gray!20, 
        colframe=gray!50, 
        boxrule=0.5pt,
        arc=3pt,
        left=5pt,
        right=5pt,
        top=5pt,
        bottom=5pt,
        sharp corners,
        fontupper=\ttfamily\tiny, 
    }
}

\input{sections/1-introduction}
\input{sections/3-system}

\input{sections/3-method}

\input{sections/4-tech_eval}

\input{sections/5-discussion}
\todo{
}

\input{sections/6-conclusion}
\input{sections/7-limitations}








\bibliography{custom}

\appendix

\clearpage

\section{Novel Insights for Anytime Index}
\label{app:novel_any}

\subsection{What does Anytime Index measure conceptually?}

The Anytime Index is a summary of a model’s entire anytime performance profile, i.e., how its expected solution quality changes as the token budget increases. This follows the classical view of anytime algorithms~\cite{zilberstein1996using}, where one characterizes an algorithm by its performance curve (quality vs. time) and often summarizes that curve via an expectation or utility over time. In our context, the Anytime Index makes explicit how quickly a method approaches high quality: whether it is “fast-thinking” (reaches high CSR early and then plateaus) or “slow-thinking” (stays weak until very large budgets). Two methods can have very similar Final CSR but quite different Anytime Index values if one achieves good quality much earlier; this is exactly the distinction we care about for judging good anytime reasoners.

In our setting: 1) The curve is CSR (or task accuracy) as a function of token budget (b) (e.g., 100-800, 200–1600, depending on the task and model); 2) Anytime Index is essentially the area under this curve, normalized to [0,1]. This is analogous to how AUC–ROC summarizes classifier performance across all thresholds rather than focusing on a single operating point. Intuitively, Anytime Index answers the question: “If I do not know in advance exactly how many tokens I will be allowed (e.g., interruption, latency budget, dynamic stopping), how good is this model on average over the whole budget range?” By contrast, Final CSR only reports quality at one point (the largest budget) and cannot distinguish methods that reach that quality quickly from those that only get there at the very end.

\subsection{Why can rankings coincide, but the metric is still useful?}

It is true that, for the experiments reported in the main table, the ordering of prompting methods and models by Anytime Index and by Final CSR is largely consistent. This outcome is not surprising: 1) The methods that perform better at the largest budget also tend to be better (or at least not much worse) at intermediate budgets; 2) The budget span on NaturalPlan is relatively narrow (e.g., 100–800 tokens), so methods that are clearly better at 800 tokens generally perform better across the entire small range as well.

\subsection{Scenario where Anytime Index provides strictly more nuanced insights than other metrics.}

While our main results emphasize that PDP improves both Final CSR and Anytime Index, the metric is particularly informative in the following concrete scenario: \textbf{Similar final CSR, different “speed” of reasoning}. Consider Figure \ref{fig:curve_grok}, both baseline CoT and PDP achieve Final CSR $\approx$ 0.95 at 800 tokens. PDP reaches 0.9 CSR already at 350 tokens and then slowly improves, while baseline CoT stays near 0.83–0.85 CSR until it jumps up at 500–600 tokens, then: Final CSR would say PDP $\approx$ baseline CoT; however, Anytime Index is higher for PDP, indicating that PDP makes Grok3-mini a better anytime reasoner under variable budgets.

We will highlight specific pairs of methods where AnyIndex reveals a larger gap than Final CSR suggests (e.g., methods whose Final CSR is close but whose early- and mid-budget behavior differs markedly). Hence, Anytime Index is not introduced to produce a different ranking in our particular tables, but to (i) align LLM-based reasoning with the classical analysis of anytime algorithms, (ii) provide a robust, budget-agnostic summary of performance over the whole budget range, and (iii) enable future work to distinguish methods that have similar final quality but very different anytime behavior.

\section{Preference Data Prompting Details}
\label{app:pref_data}
For NaturalPlan, we synthesize preference pairs for each model using the five few-shot examples provided by the dataset, which we use to construct in-context learning (ICL) prompts. Specifically, we sample $N = 64$ CoT traces per example and apply the pipeline described in Section~\ref{sec:pref_data_gen} to generate preference pairs used in Preference Data Prompting.
For AIME and GPQA, we similarly create preference data by randomly sampling 5 and 30 validation examples, respectively. For each sampled example, we generate $N = 32$ CoT traces and construct preference pairs using the same method. An illustrative preference pair from each dataset and the corresponding prompting template are provided below.


\subsection{Example Preference Pair from NaturalPlan}
\label{app:pref_data_ex}
We present example preference pairs generated by GPT-OSS-20B at the 100-token budget checkpoint. The preferred reasoning trace yields a solution with 100\% constraint satisfaction, while the rejected trace achieves 0\%, failing to account for overlapping days between travel and stay.

\subsubsection{Example task in NatualPlan}
\begin{tcolorbox}[promptbox]
You plan to visit 6 European cities for 18 days in total. You only take direct flights to commute between cities. You plan to stay in Bucharest for 5 days. From day 6 to day 10, there is a annual show you want to attend in Bucharest. You would like to visit Prague for 2 days. You want to spend 2 days in Budapest. You plan to stay in Split for 5 days. You plan to stay in Tallinn for 4 days. You have to attend a workshop in Tallinn between day 1 and day 4. You plan to stay in Dublin for 5 days. During day 10 and day 14, you have to attend a conference in Dublin.\\ Here are the cities that have direct flights:\\ Prague and Budapest, Tallinn and Prague, Bucharest and Dublin, Budapest and Bucharest, Prague and Bucharest, Prague and Split, Dublin and Split, Tallinn and Dublin, Budapest and Dublin, Prague and Dublin.\\ Find a trip plan of visiting the cities for 18 days by taking direct flights to commute between them.
\end{tcolorbox} 

\subsubsection{Preferred reasoning trace at the 100-token budget checkpoint}
Sentences contributing to effective reasoning are highlighted in green.

\begin{tcolorbox}[promptbox]
analysisWe need schedule 18 days total. Cities: Bucharest 5 days (including show day6-10). Tallinn 4 days workshop day1-4. Dublin 5 days (conference day10-14). Prague 2 days. Budapest 2 days. Split 5 days. \textbf{\textcolor{blue}{Total days sum =5+4+5+2+2+5=23 exceeds 18, so must overlap some stays? Actually you can be in city and attend events overlapping with stay counts.}} Must allocate days exactly as durations but maybe some cities visited partially? In examples \textbf{\textcolor{blue}{durations equal stay days. Here total exceeds so need ordering with overlapping constraints?}} maybe some cities can be visited same day as flight? In examples they
\end{tcolorbox}

\subsubsection{Rejected reasoning trace at the 100-token budget checkpoint}
Sentences contributing to poor reasoning are highlighted in red.

\begin{tcolorbox}[promptbox]
analysisWe need schedule 18 days total. Constraints: Bucharest 5 days total, with show day6-10 (so must be in Bucharest those days). Prague 2 days, Budapest 2 days, Split 5 days, Tallinn 4 days (workshop day1-4 must be in Tallinn). Dublin 5 days, with conference day10-14 must be in Dublin.\\ \\ We need to fit timeline. Start day1 in Tallinn for 4 days (day1-4). Then need to be in Bucharest day6-10, so must travel before day6. Also Dublin conference day10-14, overlapping with Bucharest show day6-10. Day10 both places can't. Need transition: maybe leave Bucharest day10 to Dublin arriving day10, but conference starts day10, so need be in Dublin day10. So Bucharest stay must be day5-9? Show is day6-10 inclusive, but if we leave day10 early maybe still attend? Show day6-10 inclusive, need be there through day10. Conflict. So maybe order different: go to Dublin first, conference day10-14, then Bucharest day6-10 can't after. \textbf{\textcolor{red}{Overlap impossible}}. Need overlapping days? maybe we can split
\end{tcolorbox}

\subsection{Example Preference Pair from AIME}
\label{app:pref_data_ex_aime}

We present example preference pairs generated by GPT-OSS-120B at the \textbf{1200}-token budget checkpoint. The preferred reasoning trace generates a correct solution, while the rejected trace generates an incorrect solution, imposed by a subtle but fatal logic error.

\subsubsection{Example question in AIME}
\begin{tcolorbox}[promptbox]
Let \(b \ge 2\) be an integer. Call a positive integer \(n\) \emph{\(b\)-eautiful} if it has exactly two digits when expressed in base \(b\), and these two digits sum to \(\sqrt{n}\). For example, \(81\) is \(13\)-\emph{eautiful} because
\[
81 = \underline{6}\,\underline{3}_{13}
\quad\text{and}\quad
6 + 3 = \sqrt{81}.
\]
Find the least integer \(b \ge 2\) for which there are more than ten \(b\)-\emph{eautiful} integers.
\end{tcolorbox} 

\subsubsection{Preferred reasoning trace at the 1200-token budget checkpoint}
Sentences contributing to effective reasoning are highlighted in blue.

\begin{tcolorbox}[promptbox]
analysisWe need count of n with representation two digits in base b: $n = a*b + c$ with $0\le c<b,\; 1\le a\le b-1$ (since two digits, leading digit nonzero). Condition: $a + c = \sqrt{n}$. \textbf{\textcolor{blue}{So $\sqrt{n}$ must be an integer; let $\sqrt{n} = k$. Then $n = k^2$.}} Also $n = a b + c$ with $a + c = k$. \textbf{\textcolor{blue}{So we need to find $k$ such that $k^2$ expressed in base $b$ has digits $a,c$ satisfying $a + c = k$.}} We have $a = \lfloor k^2/b \rfloor$ and $c = k^2 \bmod b$ with $1\le a\le b-1,\; 0\le c<b$.

Thus we need $k$ such that $\lfloor k^2/b \rfloor + (k^2 \bmod b) = k$. Let $k^2 = ab + c$, so $a = (k^2 - c)/b$.

Equation: $(k^2 - c)/b + c = k \Longrightarrow (k^2 + c(b-1))/b = k \Longrightarrow k^2 + c(b-1) = kb$.

Thus \textbf{\textcolor{blue}{$c(b-1) = k(b-k)$}} and so \textbf{\textcolor{blue}{$c = \frac{k(b-k)}{b-1}$}} with $0\le c<b$.

So for each $k$ we require \textbf{\textcolor{blue}{$(b-1)\mid k(b-k)$}}. Let $d = b-1$. Then $c = \frac{k(b-k)}{d}$ must be an integer, and $a = k-c$ must satisfy $1\le a\le b-1$.

Since $n=k^2<b^2$, we have $1\le k\le d$.

Thus we need $k$ such that $(b-1)\mid k(b-k)$, i.e., $d\mid k(b-k)$. Since $b=d+1$, $k(b-k)\equiv k(1-k)\pmod d$. Therefore \textbf{\textcolor{blue}{$d\mid k(1-k)\iff k(k-1)\equiv 0\pmod d$}}. \textbf{\textcolor{blue}{So the condition simplifies to $d\mid k(k-1)$}}.

Hence we count $k$ for which $k(k-1)$ is a multiple of $d$. For each prime power $p^e\mid d$, \textbf{\textcolor{blue}{$k\equiv 0$ or $1\pmod{p^e}$}}. By CRT, \textbf{\textcolor{blue}{the total number of solutions is $2^{\omega(d)}$}}, where $\omega(d)$ is the number of distinct prime factors of $d$.

We need $2^r>10$, so $r\ge 4$. The smallest $d=b-1$ with four distinct primes is \textbf{\textcolor{blue}{$2\cdot 3\cdot 5\cdot 7=210$}}, so $d=210$ and $b=211$.
\end{tcolorbox}

\subsubsection{Rejected reasoning trace at the 1200-token budget checkpoint}
Sentences contributing to poor reasoning are highlighted in red.

\begin{tcolorbox}[promptbox]
analysisWe need count numbers n with two digits in base b: $n = a*b + c$ where a,c are digits $0\le c<b$, $1\le a<b$ (since two-digit, leading digit nonzero). Sum $a + c = \sqrt{n}$. So $\sqrt{n}$ integer $\Rightarrow n$ is perfect square. Let $n = k^2$. Then $k = a + c$. Also $n = a b + c = k^2$. So we have $a b + c = (a + c)^2$.

We have $0 \le c \le b-1$, $1 \le a \le b-1$.

Let’s rearrange: $a b + c = a^2 + 2ac + c^2$ $\Rightarrow$ bring terms: $a b = a^2 + 2ac + c^2 - c = a^2 + 2ac + c(c-1)$. So

$a b = a^2 + 2ac + c(c-1)$.

Solve for b: $b = (a^2 + 2ac + c(c-1))/a = a + 2c + c(c-1)/a$.

Since b integer, \textbf{\textcolor{red}{a must divide $c(c-1)$}}. Let’s denote $d = c(c-1)$. Then $b = a + 2c + d/a$.

We need $b \ge \max(a,c)+?$ Actually b is base, must be $> \max(a,c)$, i.e., $b > \max(a,c)$. Since digits less than b, condition satisfied automatically if $b > a,c$.

We need count of $(a,c)$ producing same b? For a given b, we count solutions $(a,c)$ with $1\le a\le b-1$, $0\le c\le b-1$, and $b = a + 2c + d/a$ (integer). Equivalent to $a$ divides $d$.

Let’s brute think: For each a, \textbf{\textcolor{red}{choose $c$ where $m = c(c-1)/a$. Then $b = a+2c + m$. This determines b. So for each $(a,c)$ we get a b. We want b such that many $(a,c)$ map to same b.}} 

So need collisions. Let's compute small b maybe 2,3... but likely need larger.

We could think of equation: $b = a + 2c + c(c-1)/a$.

Define $t = c$. Then $b = a + 2t + (t(t-1))/a$.

For given b, we look for integer a dividing $t(t-1)$ and satisfying this equality.

We can attempt to find many solutions by \textbf{\textcolor{red}{choosing $t$ such that many divisors $a$ produce same $b$? But b depends on a.}} 

Probably the minimal b is 7? Let's try quick enumeration mental maybe $b=5$?

... (lengthy enumeration attempt) ...

So only one for $b=5$. Need $>10$ $\Rightarrow$ larger.

\textbf{\textcolor{red}{Probably $b=10$? Let's approximate count roughly equals number of divisors? Might be around b? Need $>10$}},
\end{tcolorbox}

\subsection{Example Preference Pair from GPQA}
\label{app:pref_data_ex_gpqa}

We present example preference pairs generated by
GPT-OSS-120B at the 800-token budget checkpoint. The preferred reasoning trace selects the correct option, while the rejected trace opts for an incorrect option, resulting from incorrect assumptions and key constraint violations. 

\subsubsection{Example question in GPQA}
\begin{tcolorbox}[promptbox]
If an equimolar mixture X of two liquids, which decolorizes bromine water, is treated with platinum when heated, then an equimolar mixture Y of two other liquids is formed as a result of disproportionation, which does not decolorize bromine water. Hydrogenation of both mixture X and mixture Y in the presence of platinum under rather severe conditions gives only one substance, a certain hydrocarbon Z (mass fraction of hydrogen is 14.28\%), which is a constituent of mixture Y and widely used as a solvent. Substance Z does not react further with hydrogen. There are no conjugated multiple bonds in the molecules of the compounds of mixture X. \\ Indicate the total number of hydrogen atoms in two liquids of mixture X.

A. 18\\ B. 22\\C. 16\\D. 12
\end{tcolorbox} 

\subsubsection{Preferred reasoning trace at the 800-token budget checkpoint}
Sentences contributing to effective reasoning are highlighted in blue.

\begin{tcolorbox}[promptbox]
analysisWe need identify X: equimolar mixture of two liquids that decolorize bromine water → they are alkenes (or phenols). No conjugated double bonds. So likely two alkenes reacting with Br2. Heated with Pt causes disproportionation to Y: mixture of two other liquids not decolorizing bromine => alkanes. Disproportionation of alkenes over Pt gives alkanes (hydrogenation) but also maybe cracking? Actually Pt at high temp causes hydrogen transfer: 2 alkenes -> one alkane more substituted, one alkene less? But result Y does not decolorize, so both are alkanes. Hydrogenation of X and Y gives same hydrocarbon Z, hydrogen mass fraction $14.28\%$ -> compute formula. Let Z be $C_n H_m$. Mass\% H = $m/(12n+m)\cdot 100 = 14.28$. Solve approx: try $n=6, m=14$? mol weight = $86$, H\% =$16.3$ too high. $n=8,m=18$: $15.8$. $n=10,m=22$: $15.5$. Need $14.28$ -> maybe $n=12,m=26$: $15.3$. lower. try $n=14,m=30$: $15.15$. Need lower. maybe $n=4,m=10$: $17.2$. Hmm. Could be benzene? $C_6H_6$: $7.7\%$ no. Maybe cyclohexane $C_6H_{12}$: $12/84 = 14.29\%$. \textbf{\textcolor{blue}{Yes! So Z = cyclohexane.}}  

So Y mixture contains cyclohexane as constituent. The other liquid perhaps also? Y is two liquids, both alkanes, one is cyclohexane, the other likely something else from disproportionation of X. Starting X are two alkenes equimolar that hydrogenate to cyclohexane → \textbf{\textcolor{blue}{both must be C$_6$H$_{10}$ isomers (cyclohexene-type).}} Two alkenes that hydrogenate to cyclohexane: could be 1-hexene and cyclohexene? But 1-hexene gives hexane, not cyclohexane. Need both give cyclohexane after Pt disproportionation? Actually Pt heated causes alkene to undergo isomerization and disproportionation: two alkenes give one alkane with longer chain and one alkene shorter. Eventually both become cyclohexane after hydrogenation. Likely X are cyclohexene ($C_6H_{10}$) and methylcyclopentene? Both $C_6H_{10}$. Two isomeric cyclohexenes? They are liquids, decolorize Br2, no conjugated. \textbf{\textcolor{blue}{Both C$_6$H$_{10}$, so total H = $10+10 = 20$. But options include 18, 22, 16, 12.}} Maybe X are cyclopentene $C_5H_8$ and 1-hexene $C_6H_{12}$ (both decolorize?). But total H =20 not listed.

Consider that after disproportionation, Y are cyclohexane ($C_6H_{12}$) and another alkane not decolorizing, maybe benzene? No.

Maybe X are two alkenes $C_3H_6$ and $C_5H_8$ totals H=14 not option.

Check options, likely 22? maybe
\end{tcolorbox}

\subsubsection{Rejected reasoning trace at the 800-token budget checkpoint}
Sentences contributing to poor reasoning are highlighted in red.

\begin{tcolorbox}[promptbox]
analysisWe need identify X: equimolar mixture of two liquids that decolorize bromine water (so they have C=C double bonds). No conjugated multiple bonds. They undergo disproportionation on Pt heated giving Y: mixture of two other liquids that does NOT decolorize bromine water (so saturated). Hydrogenation of both X and Y under severe conditions gives only one substance Z, hydrocarbon with H mass fraction $14.28\% = H\text{ mass}/M =0.1428$. Let M = molecular weight, H mass = $(n)/M =0.1428$ => $n =0.1428 M$. For hydrocarbon $C_xH_y$, $M=12x + y$. So $y/(12x+y)=0.1428$. Solve approximate: try $x=3$, $y\approx6$ giving $C_3H_6$. \textbf{\textcolor{red}{But $C_3H_6$ is unsaturated and Z “does not react further with hydrogen,” so this violates the condition.}} Next try $x=4$, giving $y\approx8$ → $C_4H_8$. \textbf{\textcolor{red}{But $C_4H_8$ is also an alkene and cannot be Z.}}  

\textbf{\textcolor{red}{Concluding Z might be C$_3$H$_6$ or C$_4$H$_8$ contradicts the requirement that Z is a fully saturated solvent.}}  

Then “Z does not react further with hydrogen” → saturated → maybe cyclohexane? They compute $C_6H_{12}$: $12/84=0.1429$ matches. So Z = cyclohexane. Good.

So Y mixture contains cyclohexane and another liquid. Disproportionation of alkenes on Pt yields alkanes? Starting X are two alkenes equimolar that decolorize bromine water, no conjugated double bonds, so likely 1-hexene and 2-hexene? But need total H atoms in two liquids of mixture X. Each molecule formula? Disproportionation (alkene coupling?) On Pt heated, alkenes undergo disproportionation: $2 RCH=CH_2 \to RCH_2-CH_2R + RCH=CH_2$? Actually yields alkane and alkene? Might produce mixture Y of two liquids not decolorizing bromine: likely cyclohexane and butane? \textbf{\textcolor{red}{But these products do not follow from the given reaction mechanism, and the chain lengths do not match the required hydrogenation product.}}  

Starting X could be 1-hexene and 2-hexene (both $C_6H_{12}$). Total H per molecule =12 each => 24 H atoms. \textbf{\textcolor{red}{But 1-hexene hydrogenates to hexane, not cyclohexane, contradicting earlier conclusions.}}  

Not match. Maybe X are $C_4H_8$ (butenes) total H = 16 total matches option C. \textbf{\textcolor{red}{But C$_4$H$_8$ hydrogenates to butane, not cyclohexane, so this is chemically impossible.}}  
\end{tcolorbox}

\subsection{Prompt used for Preference Data Prompting}
Below, we present the prompt template used for Preference Data Prompting. Each in-context example includes the task query, a preferred (good) reasoning trace, and a rejected (bad) reasoning trace. We provide one example pair for each token budget checkpoint. Below is an example prompt we used for NaturalPlan (Trip), in which case each intermediate solution quality is evaluated with CSR:

\begin{tcolorbox}[promptbox]
Here are examples showing the difference between GOOD and POOR reasoning approaches under different progress checkpoints.
You should follow the GOOD reasoning approach to reason step by step.\\

\textbf{[Scoring]}
CSR = (\# satisfied constraints) / (total constraints).\\

\textbf{[Examples @ Token Budgets]}\\
\textit{--- Pair (@ Token Budget \textbf{100} tokens) ---}\\
Question: [EXAMPLE\_QUERY]
\begin{itemize}
    \item GOOD REASONING (HIGHER CSR)[CSR=0.8]: [PREFERRED\_TRACE] Why GOOD:- Satisfaction Rate: 80.0\% - Satisfied: 8/10 constraints
    \item POOR REASONING (Lower CSR) [CSR=0.2]: [REJECTED\_TRACE] Why POOR:- Satisfaction Rate: 20.0\% - Satisfied: 2/10 constraints
\end{itemize}

\textit{--- Pair (@ Token Budget \textbf{200} tokens) ---}\\
... (omitted preference pairs at budget 200-800 tokens)\\

Now, please follow the GOOD reasoning traces in the examples to reason step by step to solve the target problem: [TARGET\_QUERY]
\end{tcolorbox}



\subsection{The computational cost of Preference Data Prompting}
\label{app:comp_cost_pdp}

\paragraph{PDP does not inherently require a large pool of reasoning traces.} 
PDP uses multiple CoT traces per example to construct preference pairs, but the number of traces (N) is a tunable hyperparameter, not a fixed requirement. In our experiments on NaturalPlan, we chose (N = 64) to capture broader variation in traces and obtain more informative preference pairs, but much smaller values of (N) (e.g., 8–16) are also possible if compute is limited. Importantly, trace generation and pair selection are one-time, offline steps per model–dataset setting. Once this pool of traces is generated and preference pairs are constructed, the same data is reused for all PDP runs and all budget settings.

\paragraph{PDP does not require running on the entire dataset or having a “complete” dataset to extract ICL examples.}
In our experiments for NaturalPlan, we rely only on the five few-shot examples provided by the benchmark and use them to extract example pairs and construct ICL-style prompts. We do not require a complete training dataset or any additional labeled data beyond these standard few-shot examples. In other domains (e.g. math reasoning), one could similarly use a small set of seed samples (e.g. randomly sample 5 samples from MATH) to generate traces and preference pairs. PDP is compatible with low-data regimes and does not depend on large-scale data.

\subsection{The connection between Anytime Index and Preference Data Prompting}
\label{app:conn_any_pdp}
Anytime Index is an evaluation construct: it summarizes the entire quality–budget curve (CSR/accuracy vs. token budget) into a single scalar, in the spirit of performance profiles for anytime algorithms or AUC over trade-off curves in classical evaluation. PDP is a training / prompting construct: it is designed to reshape that quality–budget curve, making the model achieve higher CSR earlier and more consistently across budgets.

Anytime Index effectively measures the (normalized) area under the quality-vs-budget curve over a fixed budget range. A method that: 1) achieves high CSR only at the very largest budget but stays low elsewhere will have a relatively low Anytime Index; 2) achieves moderate-to-high CSR across many budgets, especially early ones, will have a higher AnyIndex, even if the final CSR is similar. PDP is designed to favor the second type of behavior. We construct preference pairs at fixed budgets (b\_i): for each (b\_i), we compare reasoning traces that lead to higher vs. lower CSR under that same budget. The model performs in-context learning to prefer reasoning patterns that yield higher CSR at each (b\_i), not just at the maximum budget. Because we do this across all token budgets, PDP encourages trajectories that are consistently good across the whole budget range, which is precisely what leads to a larger Anytime Index.

In summary, \textbf{Anytime Index and PDP are not two unrelated contributions}: Anytime Index provides the formal objective for budget-aware anytime evaluation, and PDP is specifically designed to improve Anytime Index.

\section{Details of the Evaluation Metric of NaturalPlan (Trip)}
\label{app:no_em_trip}
In the trip planning domain in NaturalPlan, there are often many itineraries that are fully valid but do not match the single reference plan in NaturalPlan. Exact Match (EM), by construction, only counts a solution as correct if it matches the ground-truth itinerary exactly (including order, city sequence, and phrasing). In our experiments, we observe numerous cases where: the model’s itinerary satisfies 100\% of the constraints (CSR = 1.0), but differs in benign ways from the reference (e.g., a different but feasible city order, or alternative use of valid flight edges), and EM therefore assigns a score of 0. For anytime reasoning, this is problematic: two itineraries that are equally valid (CSR = 1.0) are treated as different by EM, making it hard to measure improvements in solution quality. Additionally, our core contribution is an evaluation framework for how solution quality improves as reasoning tokens increase. CSR provides a measure of partial progress (e.g., 6/10 → 8/10 → 10/10 constraints satisfied) and thus yields meaningful intermediate scores at every token budget.

\section{Constraint Satisfaction Checker Validation}
\label{app:csr_val}
We randomly sampled 100 queries from the validation set we constructed from the NaturalPlan (Trip) dataset. For each query, we used GPT-4.1 to generate a trip itinerary, resulting in 100 model outputs. Then, for each output itinerary, we ran our automatic constraint checker to compute the constraint satisfaction rate (CSR) and to explicitly list which constraints were satisfied vs. unsatisfied for each query. Next, two of the lead authors jointly inspected all 100 cases through discussion, comparing the checker’s reported CSR and satisfied/unsatisfied constraints against a careful manual assessment for each itinerary.
In 95 out of 100 cases, the checker’s output was exactly correct (both the overall CSR and the breakdown of satisfied vs. unsatisfied constraints). The remaining 5 cases involved minor discrepancies (e.g., wrong interpretations of constraints). We consider the constraint checker accurate and reliable for large-scale evaluation. 

\section{Detailed Dataset Descriptions and Hyper-parameter Configurations}
\label{app:datasets}
The NaturalPlan dataset \cite{zheng2024natural} contains three task types: trip planning, meeting planning, and calendar scheduling. We focus on the trip planning task, which poses the most complex reasoning challenges. It consists of 1,600 instances, which we split evenly into validation and test sets. The validation set is used for tuning hyperparameters, including the number of sampled CoT traces ($N$), token budget checkpoints ($b_i$), and generation length limits.

For the evaluation pipeline, we use $N=3$ to simulate diverse but efficient reasoning traces. For preference data generation, we increase the sampling to $N=64$ to capture broader trace variation. We define token budget checkpoints at $b_i \in {100, 200, 300, 400, 500, 600, 700, 800}$, aligning with typical CoT lengths across models. The maximum token limits for generating both CoT traces and solutions are set to 4,096 tokens.

We also experimented with different decoding settings and found that temperature $= 0.7$, top-$p = 1$, and top-$k = 1$ yielded the most stable results. Additionally, we incorporate the five-shot examples provided in NaturalPlan as part of the input when constructing preference data for each model.


\section{Qualitative analysis of results in NaturalPlan}
\label{app:exp_details}
\begin{figure}[h]
\centering
\begin{subfigure}{0.4\textwidth}
  \centering
  \includegraphics[width=\linewidth]{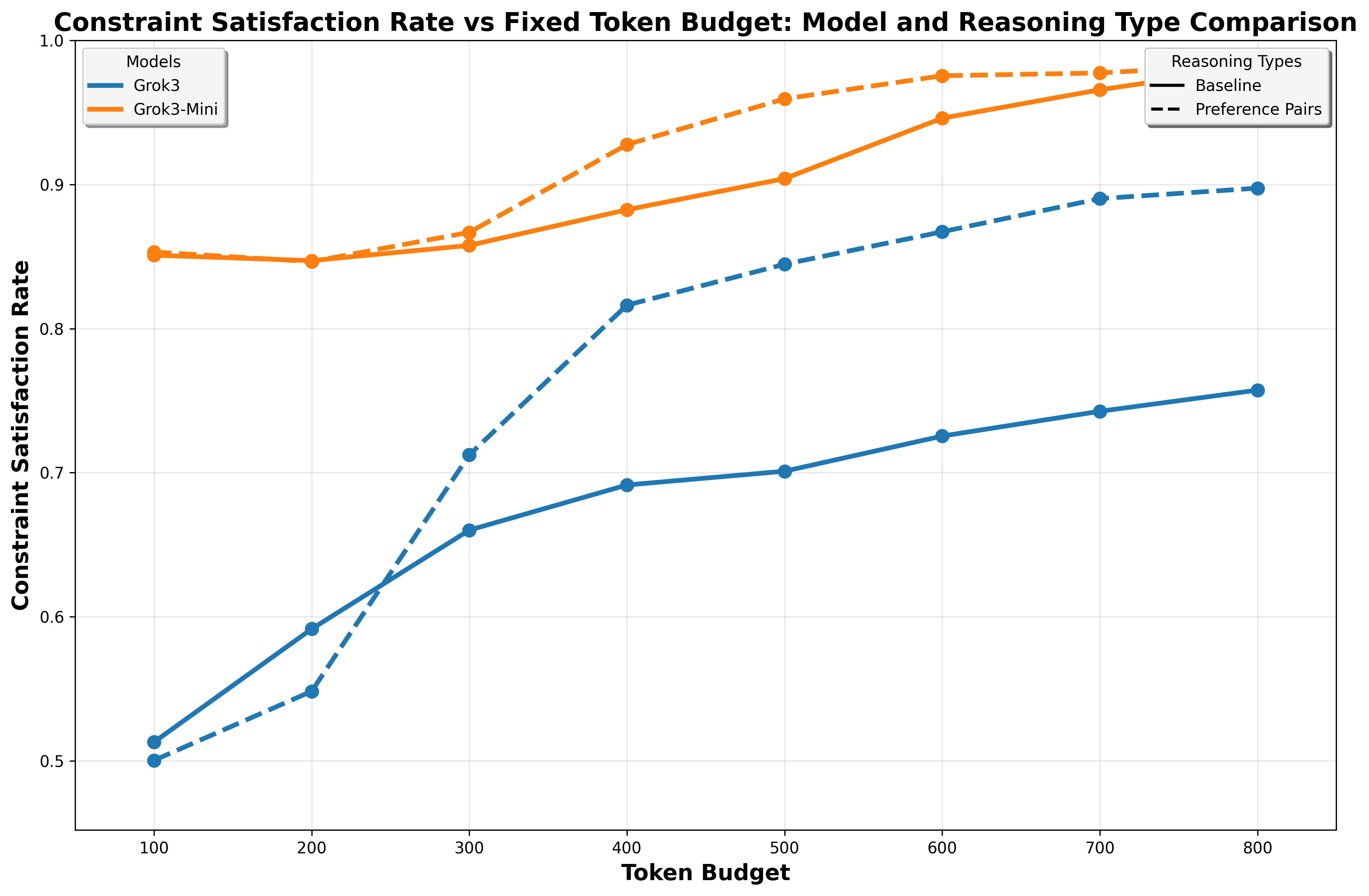}
  \caption{CSR curve for Grok-3 and Grok-3-mini.}
  \label{fig:curve_grok}
\end{subfigure}%
\hfill
\begin{subfigure}{0.4\textwidth}
  \centering
  \includegraphics[width=\linewidth]{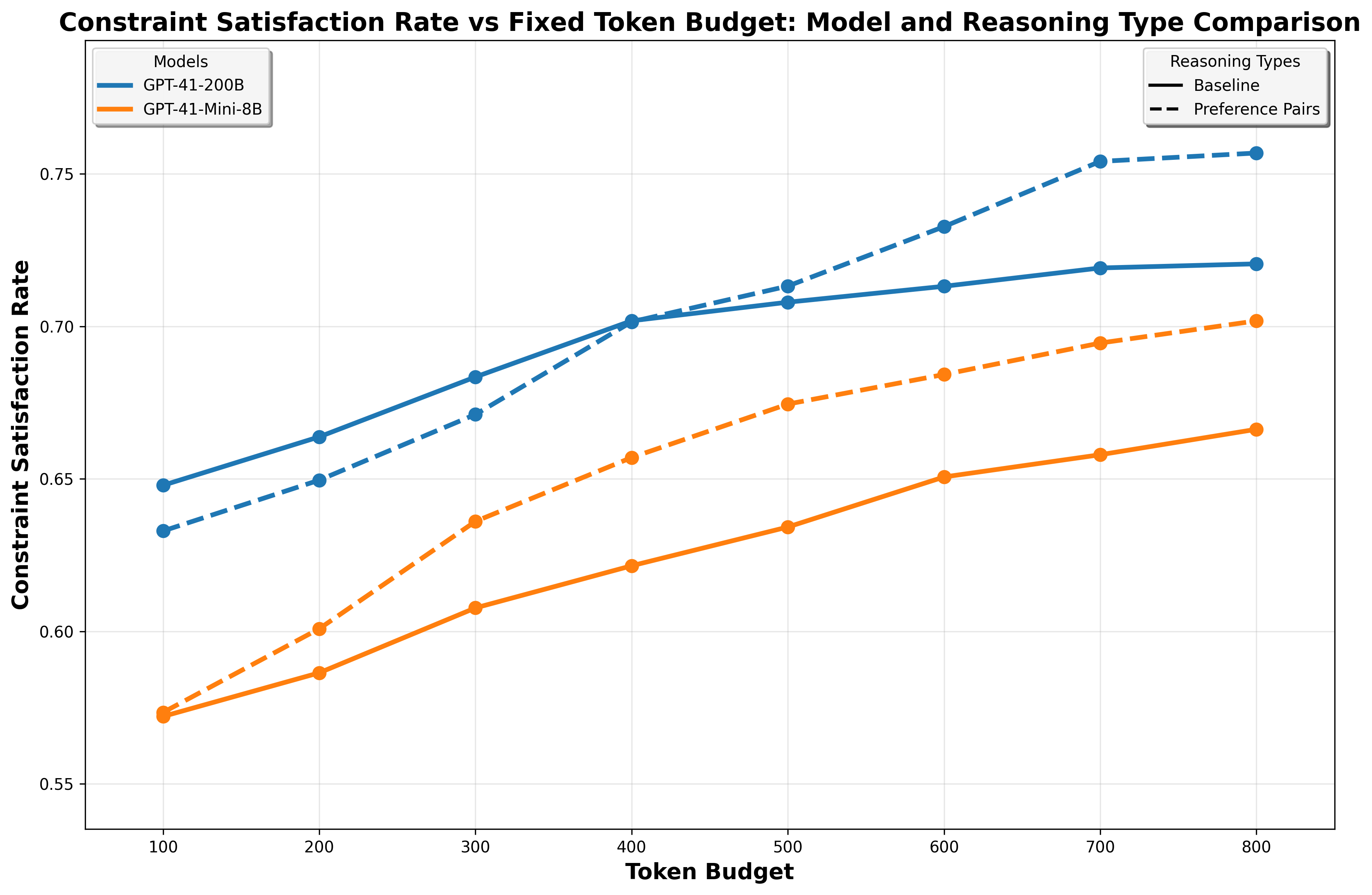}
  \caption{CSR curve for GPT-4.1 and GPT-4.1-mini.}
  \label{fig:curve_gpt41}
\end{subfigure}
\hfill
\begin{subfigure}{0.4\textwidth}
  \centering
  \includegraphics[width=\linewidth]{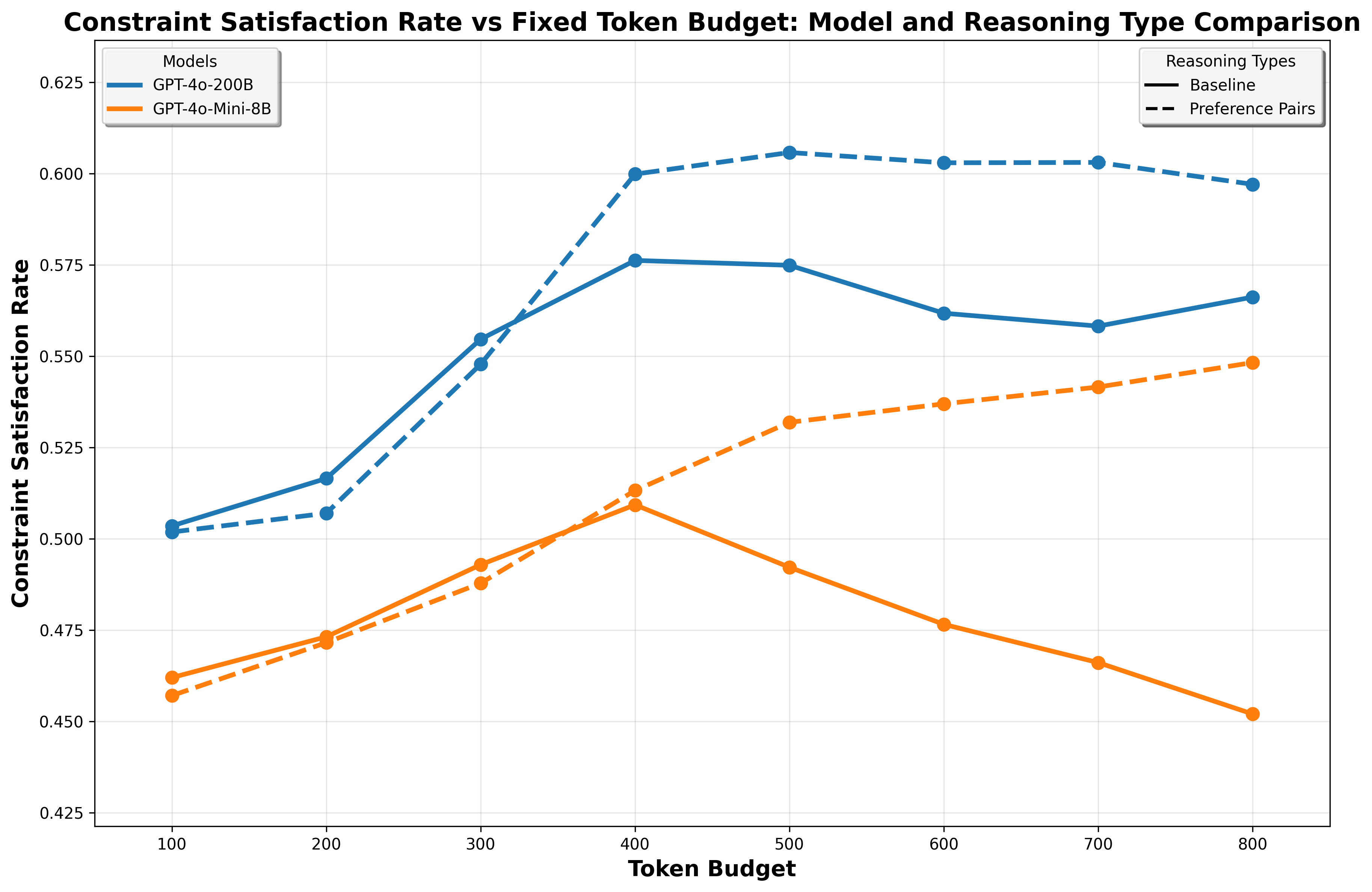}
  \caption{CSR curve for GPT-4o and GPT-4o-mini.}
  \label{fig:curve_gpt4o}
\end{subfigure}
\caption{The Constraint Satisfaction Rate evaluated at different token budget checkpoints across different model families: Grok-3, GPT-4.1, and GPT-4o. Preference Data Prompting (dotted line) makes the models better anytime reasoners.}
\label{fig:curve_q}
\end{figure}

Figure~\ref{fig:curve_q} plots constraint satisfaction rate (CSR) curves for Grok-3, GPT-4.1, and GPT-4o, comparing baseline CoT prompting with our Preference Data Prompting (PDP) approach. Across all models, PDP matches baseline performance at lower token budgets (e.g., 100, 200, 300), but begins to outperform CoT at later checkpoints. This trend underscores PDP's advantage in producing higher-quality solutions as more reasoning tokens become available.

Notably, Grok-3 benefits the most from PDP, surpassing the baseline CoT starting at the second token budget checkpoint (200 tokens). This suggests that reasoning-specialized models like Grok-3 are especially adept at leveraging preference data for self-improvement, resulting in stronger anytime reasoning performance under our PDP approach.

For GPT-4.1 and GPT-4o, the performance trend reveals a key distinction. The smaller variants of GPT-4.1 and GPT-4o exhibit continued improvement even beyond the final token budget checkpoint (800 tokens), whereas the larger models plateau. This indicates greater potential for improvement in smaller models and suggests that our approach can be particularly effective in helping them develop stronger anytime reasoning capabilities. Notably, the performance of GPT-4o shows a non-monotonic trend: as reasoning length increases, performance peaks at a medium budget and then slightly decreases. This decline is due to two main factors: (1) longer traces lead to drift and self-revision, where initial valid plans are altered and sometimes introduce inconsistencies that violate previously satisfied constraints, and (2) longer reasoning provides more opportunities for small errors that accumulate, reducing the fraction of satisfied constraints. These effects are visible in GPT-4o's CSR curve, where performance diminishes slightly at larger token budgets. Our framework is designed to surface such non-monotonic behaviors, highlighting that more reasoning is not always better and that the quality of intermediate steps is crucial for final performance.





\begin{figure}[h]
\centering
\begin{subfigure}{0.4\textwidth}
  \centering
  \includegraphics[width=\linewidth]{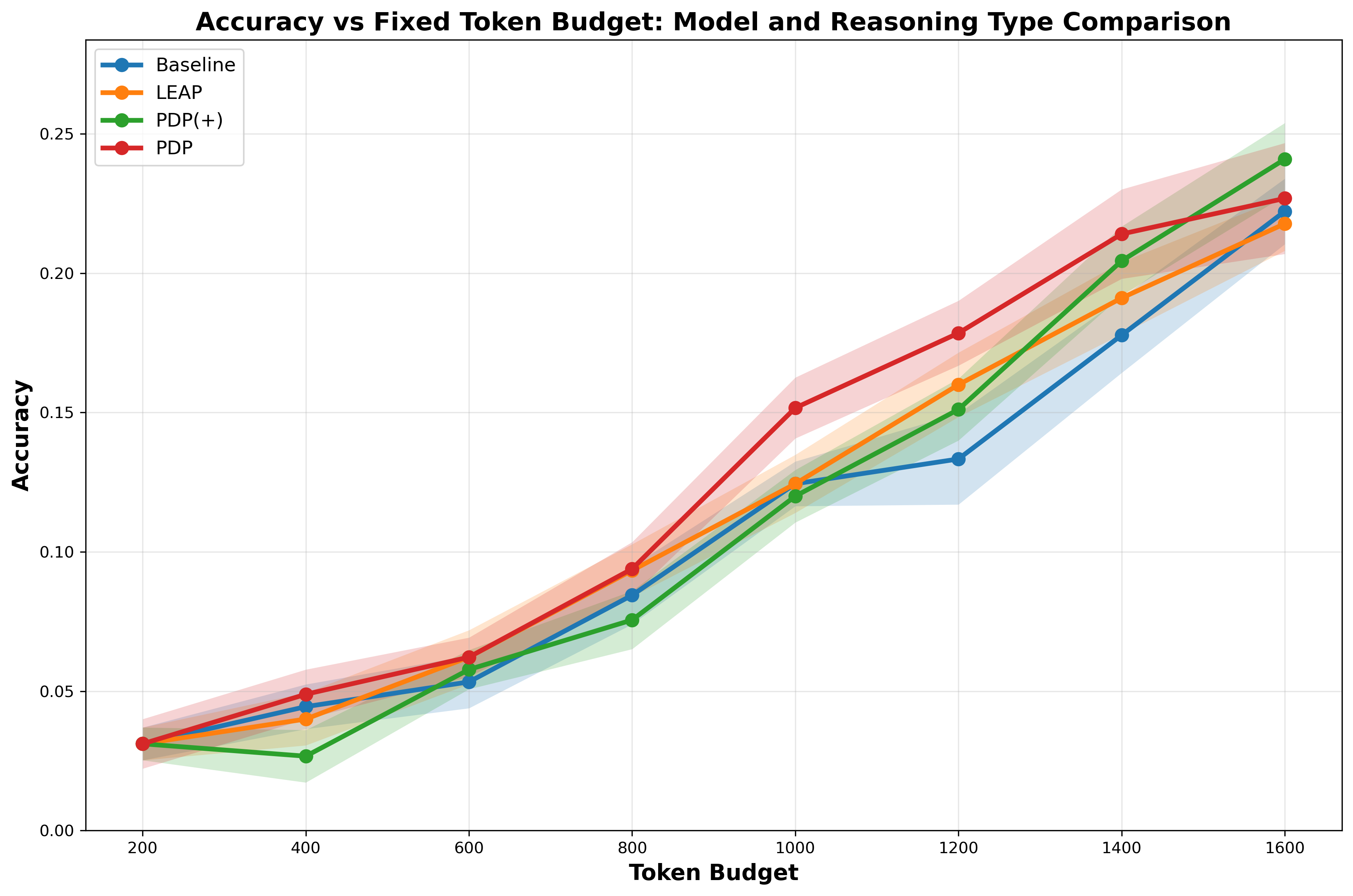}
  \caption{Accuracy curve for Grok-3 on AIME.}
  \label{fig:curve_grok3_aime}
\end{subfigure}%
\hfill
\begin{subfigure}{0.4\textwidth}
  \centering
  \includegraphics[width=\linewidth]{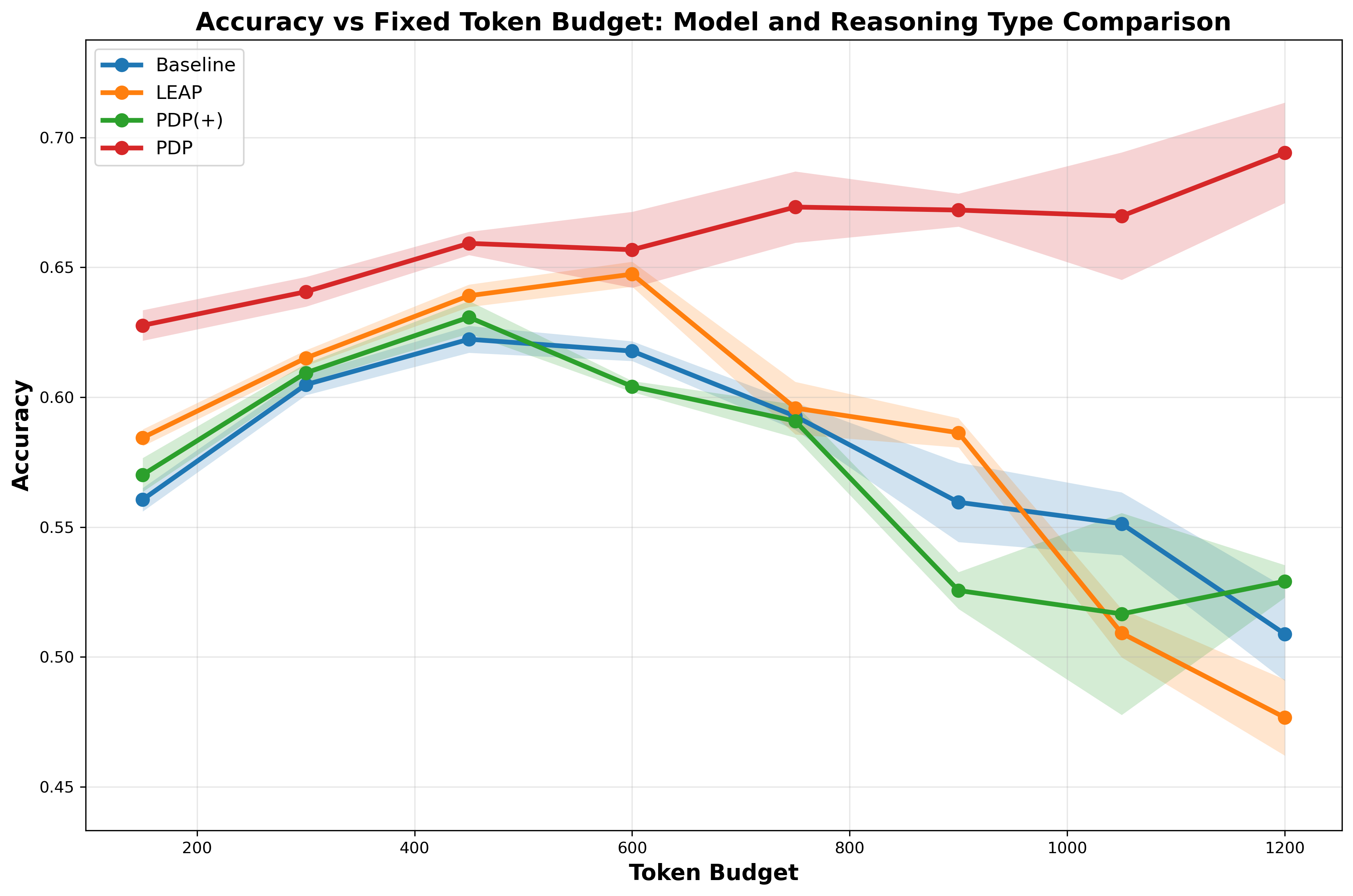}
  \caption{Accuracy curve for GPT-4.1 on GPQA.}
  \label{fig:curve_gpt41_gpqa}
\end{subfigure}
\hfill
\begin{subfigure}{0.4\textwidth}
  \centering
  \includegraphics[width=\linewidth]{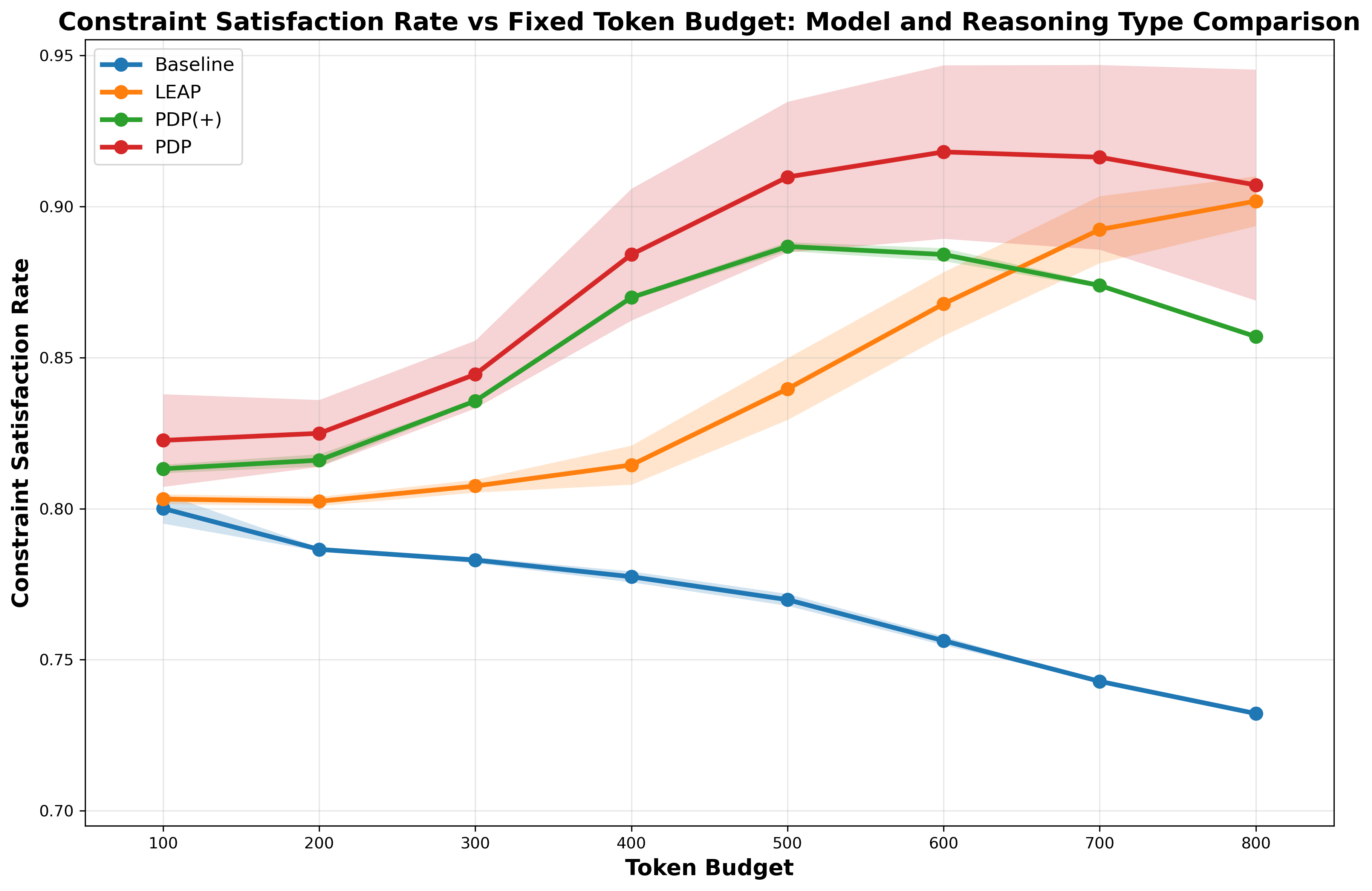}
  \caption{CSR curve for Grok-3-mini on NaturalPlan.}
  \label{fig:curve_gpt4o}
\end{subfigure}
\caption{The accuracy (on AIME and GPQA) and constraint satisfaction rate (on NaturalPlan) evaluated at different token budget checkpoints across different models: Grok-3, Grok-3-mini, and GPT-4.1. Compared to other prompting techniques, Preference Data Prompting (red line) makes the models better anytime reasoners.}
\label{fig:curve_q_four}
\end{figure}

\end{document}

%% file: sections/0-abstract.tex
We study the reasoning behavior of large language models (LLMs) under limited computation budgets. In such settings, producing useful partial solutions quickly is often more practical than exhaustive reasoning, which incurs high inference costs. Many real-world tasks, such as trip planning, require models to deliver the best possible output within a fixed reasoning budget. We introduce an anytime reasoning framework and the Anytime Index, a metric that quantifies how effectively solution quality improves as reasoning tokens increase. To further enhance efficiency, we propose an inference-time self-improvement method using LLM-synthesized preference data, where models learn from their own reasoning comparisons to produce better intermediate solutions. Experiments on NaturalPlan (Trip), AIME, and GPQA datasets show consistent gains across Grok-3, GPT-oss, GPT-4.1/4o, and LLaMA models, improving both reasoning quality and efficiency under budget constraints.

%% file: sections/1-introduction.tex
\section{Introduction}

Many real-world planning and decision-making tasks face strict compute or latency budgets. In these settings, even partial solutions can provide immediate utility (e.g., a feasible but incomplete trip plan), while additional computation can further refine them. This motivates \emph{anytime reasoning}, in which intermediate outputs improve in quality as more reasoning tokens are generated~\cite{qi2025optimizing}: a concept formalized in classic AI through anytime algorithms~\cite{zilberstein1996using, dean1988analysis, hansen2001monitoring}.


\begin{figure}[t]
\centering
        \includegraphics[width=\linewidth]{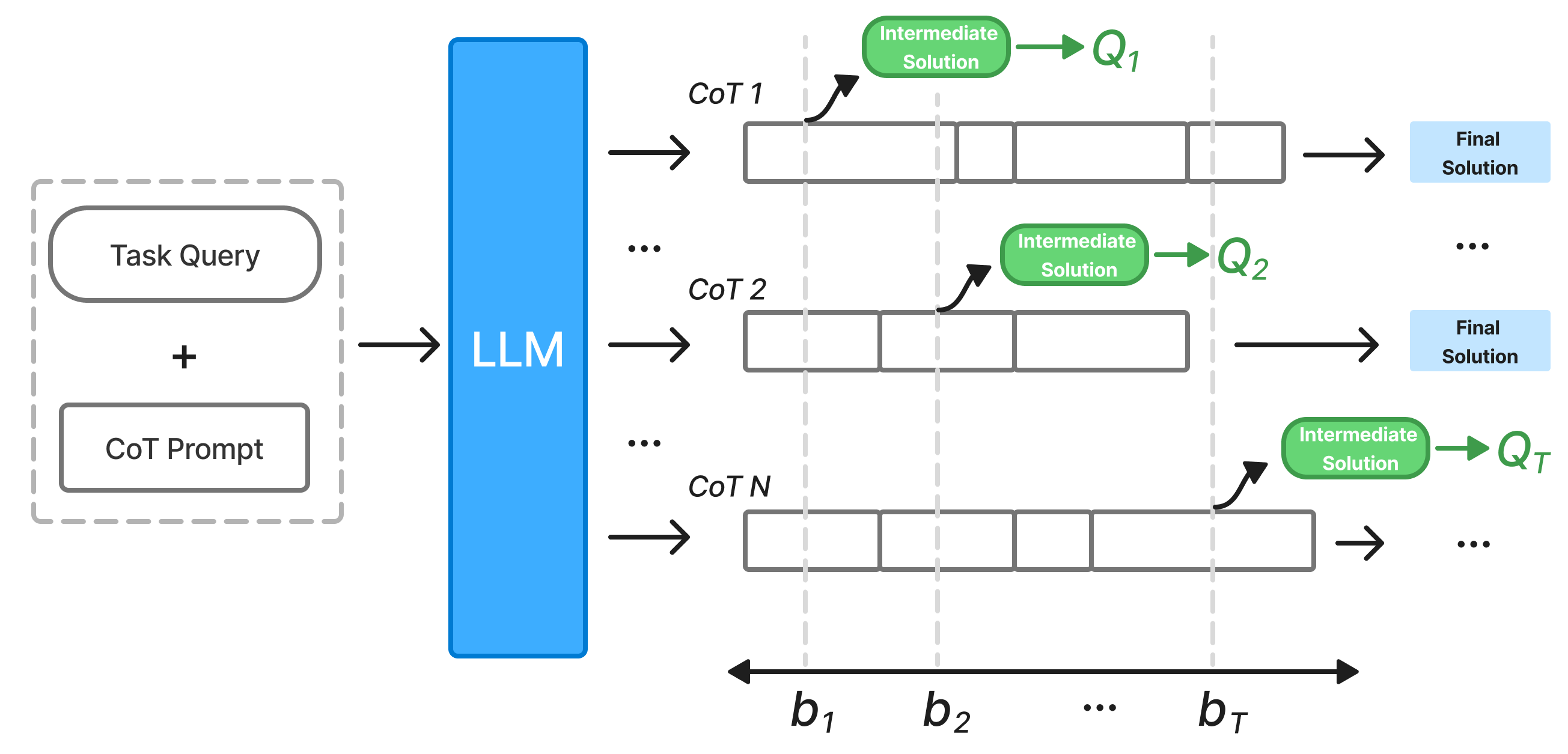}
    \caption{Overview of our anytime reasoning evaluation framework. The model generates $N$ CoT traces, each truncated at token budgets $b_i$ to evaluate intermediate solution quality $Q_i$. Final solutions are derived from full reasoning traces.}
    \label{fig:framework}
\end{figure}

Modern LLMs have shown strong reasoning capabilities through Chain-of-Thought (CoT)~\cite{wei2022chain, chowdhery2023palm}, Tree-of-Thoughts~\cite{yao2023tree}, and Self-Refine~\cite{madaan2023self}, yet these methods assume unrestricted computation and evaluate only final answer quality. Recent work has begun exploring reasoning efficiency via test-time scaling~\cite{muennighoff2025s1}, early exit strategies~\cite{yang2025dynamic}, and token-length control~\cite{han2024token}, but these approaches target final performance and offer no principled way to evaluate or improve the \emph{trajectory} of reasoning quality. Budget-aware techniques like BRPO~\cite{qi2025optimizing} and mode selectors~\cite{fang2025thinkless} optimize \emph{when} to stop thinking, but not \emph{how} to think better under constraints.


To fill this gap, we propose a framework for evaluating anytime reasoning in LLMs. We introduce the \textbf{Anytime Index}, a metric that quantifies a model's quality-per-token tradeoff across multiple reasoning budgets, capturing not just final accuracy but the efficiency of reasoning over time. We apply this framework across diverse model families, including Grok-3, GPT-oss, GPT-4.1/4o, and LLaMA-3.3, to systematically compare their reasoning efficiency under budget constraints.


To improve anytime reasoning, we further propose \textbf{Preference Data Prompting (PDP)}, an inference-time self-improvement method using LLM-generated preference data. The model generates alternative reasoning traces at fixed token budgets, identifies those leading to higher-quality intermediate solutions, and reuses the resulting preference pairs as in-context examples. We evaluate on NaturalPlan~\cite{zheng2024natural} for structured planning, AIME2024 for math reasoning, and GPQA-Diamond~\cite{rein2023gpqa} for scientific QA, showing consistent gains in both Anytime Index and task performance over strong CoT baselines.


Our contributions are three-fold:
\begin{itemize}
\item An evaluation framework for anytime reasoning in LLMs, using the Anytime Index to quantify reasoning efficiency under token budgets.
\item A scalable inference-time self-improvement method that leverages LLM-generated preference data without human supervision.
\item Empirical evidence of improved intermediate and final solution quality across structured planning, math, and scientific QA.
\end{itemize}

%% file: sections/3-system.tex
\section{Anytime Reasoning Evaluation Framework}
\label{sec:system}

\subsection{Evaluation Pipeline}
\label{sec:eval_pipeline}


The framework is illustrated in Figure~\ref{fig:framework}. For each task input, we use Chain-of-Thought (CoT) prompting~\cite{wei2022chain} to sample $N$ full reasoning traces per model, up to a global maximum of 4,096 tokens for NaturalPlan and 16,384 tokens for AIME and GPQA. Each trace includes the model’s intermediate reasoning steps and final answer. We define a series of token budget checkpoints $b_1, b_2, \dots, b_n$, where each $b_i$ corresponds to the number of reasoning tokens observed. For NaturalPlan, we use budgets $b_i \in \{100, 200, \dots, 800\}$; for AIME and GPQA, $b_i \in \{200, 300, \dots, 1600\}$.\footnote{Budget ranges are adapted to each domain's typical reasoning length.}


For each checkpoint $b_i$, we truncate the CoT trace to retain only the first $b_i$ reasoning tokens, discarding the remainder. We then re-prompt the model to generate a final answer (e.g., a trip itinerary or math solution) using only the truncated reasoning prefix as context. This simulates real-world scenarios where reasoning may be interrupted early, for instance, due to latency limits or token budgets, and the model must produce its best answer from partial computation. By freezing the reasoning at each $b_i$ and evaluating the resulting output, we can directly measure how solution quality evolves as the reasoning budget increases.

To ensure robust evaluation, we sample $N$ CoT traces per input, capturing variance from stochastic decoding. For each trace and budget $b_i$, we compute a task-specific quality score $Q_i$: Constraint Satisfaction Rate (CSR) for planning tasks, which measures the fraction of satisfied constraints, and accuracy for math and scientific QA tasks. Full metric definitions are provided in Section~\ref{sec:eval_metrics}. We report performance at each budget $b_i$ as the average quality score across all sampled traces, yielding a stable estimate of model performance across the full range of reasoning budgets.

\subsection{Anytime Index}
\label{sec:anytime_index}

To summarize a model's overall efficiency under budget constraints, we propose the \textbf{Anytime Index}, inspired by area-under-curve (AUC) metrics in optimization~\cite{qi2025optimizing}. It is defined as:
\begin{equation}
\text{Anytime Index} = \frac{\sum_{t=1}^{T-1} \frac{Q^*_t + Q^*_{t+1}}{2} \cdot (b_{t+1} - b_t)}{(b_T - b_1) \cdot Q_{\max}},
\label{eq:anytime_index}
\end{equation}
where $Q^*_t = \max_{i \leq t} Q_i$ denotes the best score achieved up to budget $b_t$, and $Q_{\max}$ is the global upper bound across all methods. The numerator computes the area under the quality curve using the trapezoidal rule, while the denominator normalizes by the budget range and the best achievable score, yielding a value in $[0, 1]$.

Intuitively, the Anytime Index captures how quickly a model approaches high-quality solutions as the token budget increases. Two methods may achieve similar final performance at the largest budget, yet differ substantially in their trajectories: a model that reaches high quality early will score higher than one that lags at smaller budgets and only improves near the end. The metric thus distinguishes ``fast-thinking'' models from ``slow-thinking'' ones under budget-aware evaluation. A detailed discussion of the additional insights provided by the Anytime Index is given in Section~\ref{app:novel_any}.

%% file: sections/3-method.tex
\section{LLM-Generated Preference Data Prompting}
\label{sec:method}

\definecolor{pdpgray}{gray}{0.92}

\begin{table*}[t]
\centering
\scriptsize
\setlength{\tabcolsep}{3.5pt}
\renewcommand{\arraystretch}{1.15}
\resizebox{\textwidth}{!}{%
\begin{tabular}{l l ccc | ccc | ccc | ccc}
\toprule
\multirow{2}{*}{\textbf{Model}} & \multirow{2}{*}{\textbf{Method}} &
\multicolumn{3}{c}{\textbf{NaturalPlan (Trip)}} &
\multicolumn{3}{c}{\textbf{AIME 2024}} &
\multicolumn{3}{c}{\textbf{GPQA}} &
\multicolumn{3}{c}{\textbf{Overall}} \\
\cmidrule(lr){3-5} \cmidrule(lr){6-8} \cmidrule(lr){9-11} \cmidrule(lr){12-14}
& &
Final & Avg & Anytime &
Final & Avg & Anytime &
Final & Avg & Anytime &
Final & Avg & Anytime \\
\midrule
\multirow{4}{*}{\textbf{Grok-3}} & Base & 74.7 & 66.8 & 68.4 & 24.0 & 11.1 & 11.0 & 69.8 & 63.5 & 63.5 & 56.2 & 47.1 & 47.6 \\
 & LEAP & 87.9 & \textbf{76.8} & \textbf{79.1} & 22.8 & 12.0 & 11.9 & 69.3 & 63.4 & 63.4 & 60.0 & {50.7} & {51.5} \\
 & PDP(+) & 89.8 & 76.6 & 78.9 & \textbf{25.0} & 11.8 & 11.5 & \textbf{70.3} & 63.7 & 63.8 & \textbf{61.7} & 50.7 & 51.4 \\ \rowcolor{pdpgray}
\cellcolor{white} 
 & PDP & \textbf{90.2} & 76.0 & 78.1 & 24.9 & \textbf{12.6} & \textbf{12.3} & 69.7 & \textbf{64.3} & \textbf{64.4} & 61.6 & \textbf{51.0} & \textbf{51.6} \\
\midrule
\multirow{4}{*}{\textbf{Grok-3-mini}} & Base & 81.5 & 76.8 & 84.7 & 80.6 & 75.2 & 80.9 & \textbf{99.3} & \textbf{90.4} & \textbf{90.6} & 87.1 & 80.8 & 85.4 \\
 & LEAP & 90.2 & 84.7 & 85.4 & 86.7 & 77.9 & 81.7 & 95.7 & 90.0 & \textbf{90.6} & 90.9 & 84.2 & 85.9 \\
 & PDP(+) & 85.7 & 85.4 & 87.6 & 83.3 & 79.0 & 82.0 & 96.9 & 85.9 & 85.8 & 88.6 & 83.4 & 85.1 \\ \rowcolor{pdpgray}
\cellcolor{white} 
 & PDP & \textbf{90.7} & \textbf{85.6} & \textbf{89.7} & \textbf{100.0} & \textbf{86.0} & \textbf{87.1} & 98.9 & 89.2 & 89.3 & \textbf{96.5} & \textbf{86.9} & \textbf{88.7} \\
\midrule
\multirow{4}{*}{\textbf{GPT-oss-120B}} & Base & 36.7 & 37.9 & 45.9 & 32.0 & \textbf{43.9} & \textbf{54.4} & 44.3 & 51.5 & {63.8} & 37.7 & 44.4 & 54.7 \\
 & LEAP & 46.1 & 38.3 & 45.2 & 30.2 & 40.5 & 49.8 & 36.8 & 47.9 & 63.6 & 37.7 & 42.2 & 52.9 \\
 & PDP(+) & \textbf{80.3} & \textbf{76.8} & \textbf{78.4} & 30.2 & 38.8 & 46.5 & 50.6 & 54.4 & 62.5 & 53.7 & 56.7 & 62.5 \\ \rowcolor{pdpgray}
\cellcolor{white} 
 & PDP & 79.5 & 76.7 & 78.3 & \textbf{38.9} & 41.7 & 52.9 & \textbf{69.4} & \textbf{66.2} & \textbf{66.2} & \textbf{62.6} & \textbf{61.5} & \textbf{65.8} \\
\midrule
\multirow{4}{*}{\textbf{GPT-oss-20B}} & Base & 51.5 & 46.4 & 47.7 & 16.5 & 22.7 & 40.4 & 28.4 & 43.5 & 58.2 & 32.1 & 37.5 & 48.8 \\
 & LEAP & 36.1 & 33.6 & 37.8 & 9.8 & 25.5 & 33.7 & 21.0 & 36.8 & 56.7 & 22.3 & 32.0 & 42.7 \\
 & PDP(+) & \textbf{58.6} & 45.3 & 47.3 & 13.9 & 28.9 & 38.6 & 34.2 & 47.3 & 58.2 & 35.6 & 40.5 & 48.0 \\ \rowcolor{pdpgray}
\cellcolor{white} 
 & PDP & 55.6 & \textbf{47.5} & \textbf{50.8} & \textbf{17.7} & \textbf{29.1} & \textbf{40.6} & \textbf{60.7} & \textbf{56.4} & \textbf{60.7} & \textbf{44.7} & \textbf{44.3} & \textbf{50.7} \\
\midrule
\multirow{4}{*}{\textbf{GPT-4.1}} & Base & 69.4 & 68.2 & 69.6 & 2.8 & 4.3 & 7.2 & 50.9 & 57.7 & 61.7 & 41.0 & 43.4 & 46.2 \\
 & LEAP & 74.6 & \textbf{70.8} & \textbf{72.2} & \textbf{10.2} & \textbf{7.5} & \textbf{9.5} & 47.7 & 58.2 & 63.7 & 44.2 & {45.5} & \textbf{48.5} \\
 & PDP(+) & \textbf{76.6} & 70.7 & 71.9 & 1.2 & 3.4 & 6.6 & 52.9 & 57.2 & 62.3 & 43.6 & 43.8 & 46.9 \\ \rowcolor{pdpgray}
\cellcolor{white} 
 & PDP & 76.3 & 70.2 & 71.4 & 7.7 & 3.9 & 5.2 & \textbf{69.4} & \textbf{66.2} & \textbf{67.1} & \textbf{51.1} & \textbf{46.8} & 47.9 \\
\midrule
\multirow{4}{*}{\textbf{GPT-4o}} & Base & 55.7 & 55.1 & 57.0 & 3.8 & 2.9 & 4.3 & 52.8 & 53.0 & {54.7} & 37.4 & 37.0 & 38.7 \\
 & LEAP & 44.1 & 49.0 & 52.4 & 3.7 & 3.0 & \textbf{5.0} & 56.4 & 52.0 & 53.1 & 34.7 & 34.7 & 36.8 \\
 & PDP(+) & \textbf{66.2} & \textbf{60.9} & \textbf{62.3} & 2.4 & 2.8 & 4.4 & 59.0 & 52.4 & 53.0 & 42.5 & 38.7 & \textbf{39.9} \\ \rowcolor{pdpgray}
\cellcolor{white} 
 & PDP & 62.8 & 58.7 & 60.2 & \textbf{5.1} & \textbf{3.1} & 4.3 & \textbf{65.3} & \textbf{54.8} & \textbf{54.9} & \textbf{44.4} & \textbf{38.9} & {39.8} \\
\midrule
\multirow{4}{*}{\textbf{Llama-3.3-70B}} & Base & 71.5 & 73.6 & 74.1 & 17.1 & 23.6 & 28.7 & 39.3 & 46.4 & 52.5 & 42.6 & 47.9 & 51.8 \\
 & LEAP & 76.2 & 74.6 & 78.4 & \textbf{33.7} & 21.7 & 26.3 & 42.8 & 44.7 & 52.0 & \textbf{50.9} & 47.0 & 52.2 \\
 & PDP(+) & 81.0 & 76.6 & 79.1 & 24.1 & 24.6 & 28.6 & 45.4 & 47.3 & \textbf{52.8} & 50.2 & 49.5 & 53.3 \\ \rowcolor{pdpgray}
\cellcolor{white} 
 & PDP & \textbf{82.0} & \textbf{78.2} & \textbf{80.2} & 22.3 & \textbf{25.3} & \textbf{29.0} & \textbf{48.1} & \textbf{47.9} & 52.4 & {50.8} & \textbf{50.5} & \textbf{53.9} \\
\bottomrule
\end{tabular}}
\caption{Experiment results across three benchmarks: trip planning (NaturalPlan), math reasoning (AIME 2024), and scientific QA (GPQA). We report final and average CSR for NaturalPlan, accuracy for AIME and GPQA, and the Anytime Index for each method. ``Overall'' columns show macro-averages across the three datasets. Best results per backbone model are in \textbf{bold}.}
\label{tab:main-results}
\end{table*}


To improve anytime reasoning under limited token budgets, we introduce a lightweight, inference-time self-improvement method that leverages LLM-generated preference data. The core idea is to teach the model to produce higher-quality intermediate reasoning traces by exposing it to contrastive examples: pairs of good and poor reasoning samples at the same token budget, generated and evaluated by the model itself. The method requires no additional supervision or fine-tuning.

Our approach is grounded in the principle that a strong anytime reasoner should maximize the Anytime Index (Section~\ref{sec:system}), reaching high-quality solutions as early as possible rather than only at the maximum budget. Inspired by recent work on the self-improvement capabilities of LLMs~\cite{huang2022large,madaan2023self,yao2024large}, we incorporate these contrastive pairs directly into the input prompt to guide the model's reasoning at inference time. We describe the preference data generation pipeline and prompting strategy below.



\subsection{Preference Data Generation}
\label{sec:pref_data_gen}

Our method builds on the evaluation setup from Section~\ref{sec:eval_pipeline}. For each task input, we sample a large pool of $N=64$ CoT reasoning traces using the base model. Each trace is truncated at each predefined token budget checkpoint $b_i$, and the model is re-prompted to generate a final answer from the truncated prefix. The resulting output is scored using task-specific metrics (CSR for planning, accuracy for math and scientific QA), yielding a quality score $Q_i$ for every trace--budget pair.

At each budget $b_i$, we rank all traces by their $Q_i$ scores and construct preference pairs: each pair consists of a preferred trace (higher $Q_i$) and a rejected trace (lower $Q_i$) truncated to the same length. Fixing the budget is critical, as it ensures that the contrast reflects differences in \emph{reasoning quality}, not reasoning length. The model learns from \emph{how} the preferred trace reasons within the same token budget, rather than simply benefiting from additional tokens. These budget-specific preference pairs form the core of our self-generated training signal, which we incorporate as in-context examples during inference (Section~\ref{sec:method}). Example preference pairs are shown in Appendix~\ref{app:pref_data_ex}.



\subsection{Preference Data Prompting}
After generating preference pairs, we incorporate them as in-context examples to guide the model's reasoning during inference. For each dataset, we select the preference pair with the largest quality gap at each token budget $b_i$, yielding one contrastive example per budget checkpoint (eight for NaturalPlan, and similarly for AIME and GPQA).

Each in-context example presents a preferred reasoning trace alongside its quality score, contrasted with a rejected trace that yields a lower-quality solution under the same budget. Crucially, we omit the intermediate solutions (e.g., the generated trip plan or math answer) from the prompt, so that the model learns from the contrastive \emph{reasoning patterns} rather than imitating surface-level outputs. By seeing what distinguishes effective reasoning from ineffective reasoning at each budget level, the model is guided toward producing higher-quality intermediate traces during inference. An example prompt is shown in Appendix~\ref{app:pref_data}. We note that Preference Data Prompting incurs minimal additional computational cost, since the preference pairs are generated once offline and reused across all inference runs. A detailed cost analysis is provided in Section~\ref{app:comp_cost_pdp}, and the connection between the Anytime Index and Preference Data Prompting is discussed in Section~\ref{app:conn_any_pdp}.

%% file: sections/4-tech_eval.tex
\section{Experiments}

\subsection{Datasets}


We evaluate anytime reasoning across three domains where solution quality can meaningfully evolve with additional computation. \textbf{NaturalPlan}~\cite{zheng2024natural} poses structured trip planning tasks, where intermediate solutions correspond to partially feasible itineraries evaluated via Constraint Satisfaction Rate (CSR). \textbf{AIME 2024} requires multi-step mathematical reasoning, where partial progress toward a solution can be assessed at each budget checkpoint. \textbf{GPQA-Diamond}~\cite{rein2023gpqa} targets expert-level scientific question answering, demanding sustained, knowledge-intensive reasoning across multiple inference steps. Together, these datasets span planning, mathematical, and scientific reasoning, providing a comprehensive testbed for budget-aware anytime evaluation. Further dataset details are provided in Appendix~\ref{app:datasets}.

\subsection{Models}
We evaluate both reasoning-specialized and general-purpose LLMs. Reasoning models include Grok-3 and Grok-3-mini~\cite{grok3_xai_2025}, GPT-oss-120B and GPT-oss-20B~\cite{openai_gpt-oss-20b_2025}. General-purpose models include GPT-4.1, GPT-4o, and Llama-3.3-70B~\cite{meta_llama3.3_70B_2024}. Each model is evaluated with four prompting strategies: (1) \textbf{Base}: Standard Chain-of-Thought (CoT) prompting~\cite{wei2022chain}; (2) \textbf{LEAP}: In-context principle learning from mistakes~\cite{zhang2024context}, which diagnoses errors on a few examples, distills explicit reasoning principles, and applies both during inference; (3) \textbf{PDP(+)}: An ablation of our method that uses only high-quality reasoning traces as in-context examples, omitting the rejected traces from preference pairs; (4) \textbf{PDP}: Our full Preference Data Prompting method, using self-generated contrastive preference pairs to guide inference-time reasoning (Section~\ref{sec:method}).

\subsection{Evaluation Metrics}
\label{sec:eval_metrics}
We use \emph{Constraint Satisfaction Rate} (CSR) for NaturalPlan\footnote{See Section~\ref{app:no_em_trip} for a discussion of why we do not use the exact-match metric proposed by \citet{zheng2024natural}.}, and standard \emph{accuracy} for AIME and GPQA.

For NaturalPlan, CSR measures the fraction of structured constraints (e.g., number of cities, total trip length, duration per city) satisfied by a model-generated itinerary. Constraints are extracted and verified automatically using a rule-based checker, validated in Section~\ref{app:csr_val}. CSR provides a graded measure of solution quality well-suited to anytime evaluation. For AIME and GPQA, accuracy evaluates whether the model's answer exactly matches the gold solution; while binary per instance, it yields a smooth signal when averaged across traces and instances.

At each token budget $b_i$, the corresponding CSR or accuracy score serves as $Q_i$ when computing the Anytime Index (Section~\ref{sec:anytime_index}). In addition to the Anytime Index, we report two complementary statistics: the final score at the maximum budget, and the average score across all budget checkpoints.

\subsection{Results}

We now evaluate two questions: whether current LLMs exhibit effective anytime reasoning behavior, and whether Preference Data Prompting (PDP) improves their reasoning efficiency under token constraints. Table~\ref{tab:main-results} reports final score, average score, and Anytime Index across NaturalPlan (Trip), AIME 2024, and GPQA for all seven models and four prompting strategies.

\paragraph{Overall performance and model-specific trends.}
Across all three benchmarks, Preference Data Prompting (PDP) consistently improves anytime reasoning performance relative to standard CoT prompting. When averaged across datasets (\emph{Overall} columns in Table~\ref{tab:main-results}), PDP achieves the highest Anytime Index for all reasoning-specialized models as well as Llama-3.3-70B. These gains are primarily driven by higher-quality intermediate reasoning traces, reflected in improved CSR/accuracy at early and mid-range token budgets. In most cases, stronger intermediate performance also carries through to better final scores at the maximum budget. 
For general-purpose models (GPT-4.1 and GPT-4o), PDP consistently improves over CoT but does not always achieve the best Anytime Index among all prompting strategies. This pattern suggests that PDP is most effective when models can reliably distinguish between higher- and lower-quality reasoning traces: a capacity that is stronger in reasoning-specialized models. This observation aligns with prior work showing that stronger reasoners benefit more from self-generated preference signals~\cite{pang2024iterative, chen2024learning}. Example performance curves illustrating these trends are shown in Figure~\ref{fig:curve_q_four}.


\paragraph{The role of contrastive preference supervision.}
To isolate the effect of contrastive supervision, we compare PDP against PDP(+), an ablation that uses only high-quality reasoning traces as in-context examples. PDP(+) consistently improves over CoT, confirming that exposure to strong reasoning trajectories is beneficial on its own. However, the full PDP method, which includes both preferred and rejected traces, yields stronger anytime behavior. Across the 21 model--dataset settings, PDP achieves a higher Anytime Index than PDP(+) in the majority of cases and also improves average CSR/accuracy more consistently. The gap is most pronounced for reasoning-specialized models, where PDP substantially outperforms PDP(+) on both average scores and Anytime Index. These results demonstrate that the contrastive signal, showing the model \emph{what to avoid}, not just what to emulate, provides meaningful additional guidance beyond exposure to good examples alone.


\paragraph{Comparison with learning-from-mistakes baselines.}
We also compare PDP against LEAP~\cite{zhang2024context}, a strong baseline that learns explicit principles from model-generated mistakes. While LEAP often outperforms vanilla CoT, particularly on NaturalPlan and AIME, PDP consistently matches or surpasses LEAP on the Anytime Index. As shown in Table~\ref{tab:main-results}, PDP achieves a higher \textit{overall} Anytime Index than LEAP across six out of seven models, and it also demonstrates superior average CSR/accuracy in all cases. This advantage is most notable for models with strong reasoning capabilities, where PDP improves \textit{overall} final CSR/accuracy by 17.6\%, average CSR/accuracy by 11.4\%, and the Anytime Index by 7.9. These results highlight that preference-based contrastive supervision more effectively guides intermediate reasoning under limited token budgets, outperforming in-context learning from principles derived from model-generated mistakes.

%% file: sections/6-conclusion.tex
\section{Conclusion}

We introduced the Anytime Index, a metric for evaluating quality-per-token efficiency in LLM reasoning under budget constraints, and proposed Preference Data Prompting, an inference-time self-improvement method that uses LLM-synthesized preference data to guide reasoning toward higher-quality intermediate outputs without human supervision. Experiments on structured planning, math reasoning, and scientific QA show consistent gains in both task performance and Anytime Index across model families. In future work, we plan to investigate how anytime reasoning behavior can be internalized during training through preference-driven fine-tuning.




%% file: sections/7-limitations.tex
\section{Limitations}
\label{sec:limitations}

While our work introduces a novel framework for evaluating and improving anytime reasoning in LLMs, several limitations remain. First, our experiments focus primarily on the trip planning domain using the NaturalPlan dataset. Although this domain is well-suited for evaluating structured, constraint-based reasoning under budget constraints, future work should validate the framework on a broader set of tasks, including open-ended domains such as code generation.

Second, we primarily compare standard Chain-of-Thought prompting with our proposed preference data prompting. While sufficient for demonstrating the utility of the Anytime Index, we do not exhaustively benchmark against other prompting strategies (e.g., Tree-of-Thoughts, Self-Consistency, etc.) Expanding our evaluation to include a wider range of reasoning methods would provide deeper insights into the relative strengths of different approaches under budget constraints.

Finally, our self-improvement method is limited to inference-time prompting with LLM-generated preference pairs. While effective, this leaves open the opportunity to train models explicitly for better anytime behavior using preference-driven fine-tuning methods such as Direct Preference Optimization (DPO). If prompting already yields meaningful gains, integrating preference supervision into the training loop could further improve budget-aware reasoning and close the self-improvement cycle.